\documentclass[letterpaper, 10 pt, conference]{ieeeconf}  

\IEEEoverridecommandlockouts                              

\usepackage{times} 

\usepackage{cite}
\usepackage{amsmath,amssymb,amsfonts}
\usepackage{algorithmic}
\usepackage{graphicx}
\usepackage{subfig}
\usepackage{textcomp}
\usepackage{xcolor}

\title{\LARGE \bf
A machine learning framework for acoustic reflector mapping
}

\author{Usama Saqib$^{1}$, Letizia Marchegiani$^{2}$, and Jesper Rindom Jensen $^{3}$
\thanks{*This work was not supported by any organization}
\thanks{$^{1}$Author is an independent researcher based in Denmark.
        {\tt\small usamasaqib@gmail.com}}%
\thanks{$^{2}$Authors are with Department of Engineering and Architecture, University of Parma, Italy
        {\tt\small letizia.marchegiani@unipr.it}}%
\thanks{$^{3}$Author are with Audio Analysis Lab, Department of Electrical Systems, Aalborg University, Denmark
        {\tt\small jrj@es.aau.dk}}%
}

\begin{document}

\maketitle
\thispagestyle{empty}
\pagestyle{empty}

\begin{abstract}
Sonar-based indoor mapping systems have been widely employed in robotics for several decades. While such systems are still the mainstream in underwater and pipe inspection settings, the vulnerability to noise reduced, over time, their general widespread usage in favour of other modalities(\textit{e.g.}, cameras, lidars), whose technologies were encountering, instead, extraordinary advancements. Nevertheless, mapping physical environments using acoustic signals and echolocation can bring significant benefits to robot navigation in adverse scenarios, thanks to their complementary characteristics compared to other sensors.  Cameras and lidars, indeed, struggle in harsh weather conditions, when dealing with lack of illumination, or with non-reflective walls. Yet, for acoustic sensors to be able to generate accurate maps, noise has to be properly and effectively handled. Traditional signal processing techniques are not always a solution in those cases. In this paper, we propose a framework where machine learning is exploited to aid more traditional signal processing methods to cope with background noise, by removing outliers and artefacts from the generated maps using acoustic sensors. Our goal is to demonstrate that the performance of traditional echolocation mapping techniques can be greatly enhanced, even in particularly noisy conditions, facilitating the employment of acoustic sensors in state-of-the-art multi-modal robot navigation systems. Our simulated evaluation demonstrates that the system can reliably operate at an SNR of $-10$dB. Moreover, we also show that the proposed method is capable of operating in different reverberate environments. In this paper, we also use the proposed method to map the outline of a simulated room using a robotic platform. 
\end{abstract}

\section{INTRODUCTION}
\label{sec:intro}

Simultaneous localization and mapping (SLAM) algorithms predominately rely on sensors such as cameras and lasers to build maps of the environment and localise against those maps. Yet, such systems are vulnerable to specific conditions: \textit{e.g.}, cameras struggle in case of low visibility, and lidars suffer in harsh weather. Furthermore, these technologies encounter difficulties when handling reflective objects and surfaces \cite{yang2020pi}. This limits their applicability in constructing spatial maps of certain indoor environments, \textit{e.g.}, store centres, where glass doors and shop windows are ubiquitous.

Echolocation has been extensively studied in the past, leading to the development of acoustics-based mapping systems \cite{eliakim2018, kuc2012echolocation}. Such systems are not affected by the environment's appearance, weather or lighting conditions, but their performance can decrease abruptly in the presence of noise \cite{yang2020pi, mo2023survey}. Acoustic sensors have, indeed, complementary characteristics compared to the above-mentioned sensors, which make them a precious asset in the development of robust and reliable multi-modal mapping systems. Additionally, acoustics-based mapping modules are computationally much lighter than vision and laser-based ones, making them particularly convenient for resource-constrained robots (\textit{e.g.}, drones) \cite{dumbgen2022blind}. In particular, we opted for the use of audible signals and microphones, as, compared to other proximity sensors (\textit{e.g.} ultrasound, infrared), has some benefits when operating in the specific settings we are interested in. Firstly, the variety of polar patterns of microphones allows the use of less directional ones, reducing the number of sensors necessary to cover the entire environment; secondly, robots which are used in malls-like contexts are generally also designed to interact with users and are factory-equipped with microphones for that purpose. Thus, it would be advantageous, both for resource-constrained robots and interactive ones to rely on audible signals also for navigation tasks.

Mapping using echolocation essentially entails identifying and localizing the surrounding acoustic reflectors. In literature, this is generally achieved through the estimation of the \textit{time of arrival} (TOA) and the \textit{direction of arrival} (DOA) of the reflected echoes from the source to the microphone, as the robot moves around the environment (\textit{e.g.}, \cite{Nguyen2019, fan2020acoustic}). Such a process can be particularly challenging, though, as it requires modelling the different environmental factors which can affect the localization of acoustic reflectors. Furthermore, one noticeable problem encountered when using TOA and DOA estimators is that they will provide an estimate no matter where the robot is located in an environment. Hence, those systems would not distinguish between an estimate actually belonging to an acoustic reflector from one belonging to an empty space.

In our previous work \cite{saqib2019, jensen2019, saqib2020estimation, saqib2022}, we provide solutions to estimate TOA and DOA directly from the observed signal, independently of the specific environment. In this paper, building upon our previous efforts, we propose a framework that addresses all the above-mentioned issues and extends the state-of-the-art by:
\begin{itemize}
        \item introducing a novel and computationally lighter method to obtain the TOA and DOA estimates which uses an UCA to acquire raw acoustic signals which is later passed to a TOA estimator based on Non-linear least squares (NLS) and a DOA estimator based on beamforming techniques.
        \item filtering such estimates applying an SVM-based classifier able to discriminate between an acoustic echo and empty space;
        \item performing a thorough analysis of the behaviour of our system at different SNR levels, to demonstrate its resilience to noise, which is considered one of the major drawbacks of acoustic-based mapping systems.
\end{itemize}

We evaluate our framework by evaluating the proposed method within a simulated environment as it gives us flexibility to control the signal-to-noise ratio (SNR), and reverberation (T60).

The structure of the paper is as follows: Section \ref{sec: signalModel} contains the signal model and the problem formulation. In Sections \ref{sec:proposedDoaToaEst}, \ref{sec: doaEstimator} and \ref{sec:trainingSvm}, we introduce the proposed TOA, DOA estimator and classifier, respectively. Finally, the experimental results as well as the discussion and conclusion are found in section \ref{sec: expResults} and \ref{sec: conclusion}, respectively.

\section{Related Works}
\label{sec: related_works}


Current state-of-the-art techniques have addressed the problem of estimating the acoustic reflector of an environment using echolocation. For instance, the authors in \cite{echoSLAM2016, Nguyen2019, boutin2020drone} have proposed several methods to infer distance to an acoustic reflector as the robot moves within an environment. The authors assume that the time of arrival (TOA) knowledge is known and could be extracted from the Room Impulse Response (RIR) of an environment using a standard peak-picking approach. However, TOA estimation from the estimated RIR is non-trivial in practice \cite{kelly2014} and the individual RIRs need to be estimated as the robot moves within an environment. However, TOA estimation alone does not help in the construction of a spatial map of an environment. Along with TOAs, the knowledge about the direction-of-arrival (DOA) of the echoes is also required. DOA estimation is found in several research papers \cite{valin2004, pan2014, sun2011}. Recent advancement in Machine Learning techniques especially Deep Learning for DOA estimation can be seen in \ref{grumiaux2022survey, chen2022multiple, pavel2021machine}. In \ref{pavel2021machine}, the authors are trained a neural network to teach the relationship between the input sparse covariance matrix and the true signal directions. 


\section{Problem Formulation}
\label{sec: signalModel}
Let us consider a setup with a mobile robot equipped with both a uniform circular array with $M$ microphones, located on top of it, and an omnidirectional loudspeaker, for probing the environment, located at the center of the array, as seen in our earlier work \cite{saqib2022}. The loudspeaker emits a known signal $s(n)$ which is recorded by the microphones $m = 1,\dots, M$. 
The observed signal recorded by microphones $y_{m}(n)$ is then modeled as follows:
\begin{align}
\label{eq:signalModel}
y_{m}(n) &= (h_{m}*s)(n) + v_{m}(n)
\end{align}
where $v_{m}(n)$ is the combined interfering source, \textit{e.g.}, the ego-noise of the robot, and the background noise plus the sensor noise. Moreover, $h_{m}(n)$ is the RIR from the loudspeaker to microphone $m$ and $*$ represents the convolution operator. Here, $s(n)$ is assumed to be known.
To facilitate the estimation of the TOA and DOA, we could rewrite (\ref{eq:signalModel}) as the sum of the first $R$ reflections in noise given the structure of the RIR \cite{saqib2020estimation}, \textit{i.e.},
\begin{align}
    \label{eq:sumofAllReflections}
    y_{m}(n) & = \sum_{r=1}^{R} \notag g_{m,r}s(n-\tau_{ref,r}-\eta_{m,r}) + u_{m}(n)\\
    & = x_{m}(n) + u_{m}(n)
\end{align}
where $g_{m,r}$ is the gain or attenuation of the $r$th reflection from the loudspeaker to microphone $m$, and $R$ is the number of early reflections including the direct-path component. Furthermore, $\eta_{m,r} = \tau_{m,r} - \tau_{ref,r}$ is the time difference of arrival (TDOA) of the $r$th component measured between the reference point and microphone $m$, $\tau_{ref,r}$ is the TOA of the $r$th component measured between reference point and the acoustic reflector position and $\tau_{m,r}$ is the TOA between microphone $m$ and the acoustic reflector, while $u_{m}(n)$ is the noise term including interfering source, the background plus sensor noise, and late reflections. In our definition, the direct-path component of the probe sound, $s(n)$, is the component corresponding to $r=1$. Moreover, microphone $1$ is chosen as the reference point. Assuming that the reflectors are in the far-field of the array and given the geometry of the microphones and loudspeaker, we can write the TDOA, $\eta_{m,r}$, of the UCA as follows:
\begin{align}
\label{eq:3}
    \eta_{m,r}=r\sin\psi_r[\cos(\theta_1-\phi_r)-\cos(\theta_m-\phi_r)]\frac{f_s}{c},
\end{align}
where $\psi$ and $\phi$ are the elevation and azimuth angle, respectively, while $\theta_{1}$ is the angle of the reference microphone and $\theta_{m}$ is the angle of the remaining microphones for $m = 1, \dots, M$, respectively. Moreover, $f_{s}$ is the sampling frequency and $c$ is the speed of sound. The reference microphone angle $\theta_{1}$ is defined as
\begin{align}
\label{eq:4}
\theta_1=\phi_r+\alpha,
\end{align}
where $\alpha$ is the offset angle. The $m$'th microphone $\theta_{m}$ is defined as
\begin{align}
\label{eq:5}
    \theta_m=\theta_1+\frac{2\pi(m-1)}{M},
\end{align} If we collect $N$ time samples from each microphone and assume stationarity across those samples, we can vectorize our data and extend our signal model as 
\begin{align}
\mathbf{y}_{m}(n)&= \mathbf{x}_m(n)+\mathbf{u}_m(n)\\
&=\begin{bmatrix}y_{m}(n)&\cdots&y_{m}(n+N-1)\end{bmatrix}^T,
\end{align}
where the time-stacked probe signal, $\mathbf{x}_m(n)$, and noise, $\mathbf{u}_m(n)$, are defined similarly to $\mathbf{y}_m(n)$. The objective is then twofold. Firstly, we need to estimate the unknown TOAs and DOAs of the $R$ early reflections from the $N$ time samples from each of the $M$ microphones. Secondly, we need to detect whether the obtained estimates belong to an actual acoustic reflector, or they are spurious ones due to poor estimation conditions.

\subsection{Sequential Non-linear Least Squares (S-NLS)}
\label{sec:proposedDoaToaEst}
If $N$ samples of the reflected signals are taken while assuming that $s(n)$ is known and the robot position is assumed fixed within these $N$ samples, then a nonlinear least squares (NLS) estimator can be formulated, which is statistically optimal under white Gaussian noise conditions. This is expressed as follows:
\begin{align}
\label{eq:7}
\{\widehat{\mathbf{g}},\widehat{\boldsymbol{\tau}}\} 
&=\operatorname*{arg\,min}_{\mathbf{g},\boldsymbol{\tau}}\left\| \mathbf{y}_{1}(n)-\sum_{r=1}^{R} g_{r}s(n-\tau_{r})\right\|^2,\\
 \boldsymbol{\tau} &= \begin{bmatrix}{\tau}_{1}& {{\tau}}_{2}& \cdots &{\tau}_{R}\end{bmatrix}^T,\\
\mathbf{g} &= \begin{bmatrix}{g}_{1}& {g}_{2}& \cdots &{g}_{R}\end{bmatrix}^T,
\end{align}
$\mathbf{\tau}$ and $\mathbf{g}$ are the TOAs and attenuation of the $R$-th reflections and $\widehat{\tau}$ and $\widehat{g}$ denotes an estimate of $\tau$ and $g$. Since only one microphone is chosen for the TOA estimate,  we can reduce the computational load of the cost function \eqref{eq:7} compared to the work presented in \cite{saqib2022}. To solve \eqref{eq:7}, we first convert all the observations into the frequency domain to make the processing computationally efficient. The signal model from the reference microphone $1$ in \eqref{eq:sumofAllReflections} is:
\begin{align}
    \label{eq:9}
    \mathbf{Y}_{1}(\omega)&= \mathbf{X}_{1}(\omega)+\mathbf{U}_{1}(\omega),
\end{align}
where the $\mathbf{Y}_1(\omega),  \mathbf{X}_{1}(\omega)$ and $\mathbf{U}_{1}(\omega)$ are the Fourier transforms of $y_1(n), x_1(n)$ and $u_1(n)$, respectively. The signal model in \eqref{eq:7} in frequency domain is written as:
\begin{align}
\label{eq:10}
    \{\widehat{\mathbf{g}},\widehat{\boldsymbol{\tau}}\} 
&=\operatorname*{arg\,min}_{\mathbf{g},\boldsymbol{\tau}}\left\| \mathbf{Y}_1-\sum_{r=2}^Rg_r\mathbf{Z}(\tau_r)\odot \mathbf{S})\right\|^2,\\
    \mathbf{Z}(\tau)&=\begin{bmatrix}1& e^{-j\tau2\pi\frac{1}{K}} & \cdots & e^{-j\tau2\pi\frac{K-1}{K}} \end{bmatrix}^T.
\end{align}
Here, the observation $\mathbf{Y}_1$ is taken from the microphone $1$, and the frequency index is omitted for a simpler notation. The vector, $\mathbf{Z}(\tau)$, delays the source signal $\mathbf{S}$ by $\tau$ samples through a phase shift, while $\odot$ is the element-wise product operator. In order to estimate the gain and TOA parameters of the multiple reflections, $R$, various cyclic methods could be used like the RELAX method proposed in \cite{jian1996} that iteratively estimates the values of $\boldsymbol{\tau}$ and $\mathbf{g}$. However, the estimator can be simplified since it is linear with respect to the unknown gains. In the special case, for example, if we are interested in estimating only the strongest echo, then we can set $R=1$. Therefore, we can solve \eqref{eq:10} for $g$ by taking the derivative of the cost function and equating to zero yields
\begin{align}
\label{eq:11}
\widehat{g}_{r} = \frac{\mathbf{{Y}}_1^H\mathbf{\overline{Z}}(\tau_r)+\mathbf{\overline{Z}}^H(\tau_r)\mathbf{{Y}}_1}{2\mathbf{\overline{Z}}^H(\tau_r)\mathbf{\overline{Z}}(\tau_r)}
\end{align}
where $\overline{\mathbf{Z}}(\tau)=\mathbf{Z}(\tau) \odot \mathbf{S}$. This can be inserted back into the estimator in \eqref{eq:10} to estimate $\widehat{\tau}$ and simplify for single reflection to get 
\begin{align}
\label{eq:13}
& \widehat{\tau}= \operatorname*{arg\,max}_{\tau}{\rm I\!R} \{ \mathbf{Y}_{1}^{H}\overline{\mathbf{Z}}(\tau)\}
\end{align}
where the operator $\rm I\!R$ represents taking the real part of the vector. 

\subsection{DOA Estimation}
\label{sec: doaEstimator}
Although any kind of spatial filter could be applied to the proposed framework, in this paper, we use a minimum power distortionless response (MPDR) beamforming on the microphone $1$ observation:
\begin{align}
    {U}(\omega)&= \mathbf{w}^H(\omega){\mathbf{Y}_{1}}
\end{align}
where $U$ is the filtered observed signal processed by the beamformer, $\mathbf{w}$. The objective of the beamformer is to recover the clean signal at the reference microphone from the noisy observations. The MPDR solution to this problem can be shown to equal
\begin{equation}
    \label{eq:15}
    \mathbf{w}_{\text{MPDR}} = \frac{\mathbf{R}^{-1}_{y}\mathbf{d}(\psi, \phi)}{\mathbf{d}^{H}(\psi, \phi)\mathbf{R}^{-1}_{y}\mathbf{d}(\psi, \phi)},
\end{equation}
where $\mathbf{d}^{H}(\psi, \phi)$ is the steering vector, $\psi$ and $\phi$ denotes the steering angles, elevation and azimult, respectively, and $\mathbf{R}_y=E[\overline{\mathbf{Y}}\overline{\mathbf{Y}}^H]$ is the $M\times M$ covariance matrix of the observed signal. The DOA of the reflected signal can then be estimated using a steered response power approach by maximizing the output power of the MPDR beamformer versus the steering angle. That is,
\begin{align}
\label{eq:16}
\{\widehat{\psi},\widehat{\phi}\}&=\arg\max_{\psi}UU^H\\
&=\arg\max_{\psi,\phi}\mathbf{w}^H\mathbf{R}_y\mathbf{w},
\vspace{-0.4cm}
\end{align}

\subsection{Acoustic Echoes Classifier}
To classify an acoustic reflector from a non-reflector,  \textit{i.e.}, we rely on a Radial Basis
Function (RBF)-based SVM classifier,  operating on two features: the delay of acoustic echo, \textit{cf.} Eq. \eqref{eq:13}, and the output power of the beamformer, \textit{cf.} Eq. \eqref{eq:16}.

\section{Implementation Details}

\subsection{The Beamformer}
To implement the beamformer, we use the overlap-add technique \cite{pan2014}. The output of the microphone was divided into overlapping frames with a frame width of $882$ samples ($20$ ms with a sampling rate of $22.05$ kHz) with a window overlap of $50~\%$. Later, each frame is multiplied with a Hanning window. These frames are then transformed using a Short-Time Fourier Transform (STFT). For each frequency bin, a beamformer was designed and applied to the received signals $\mathbf{Y}(t,\omega)$ for time frame $t$. Furthermore, for each sub-band, the observed signal covariance matrix is estimated as
\begin{figure*}[t]
    \centering
    \subfloat[]{{\includegraphics[scale=0.5]{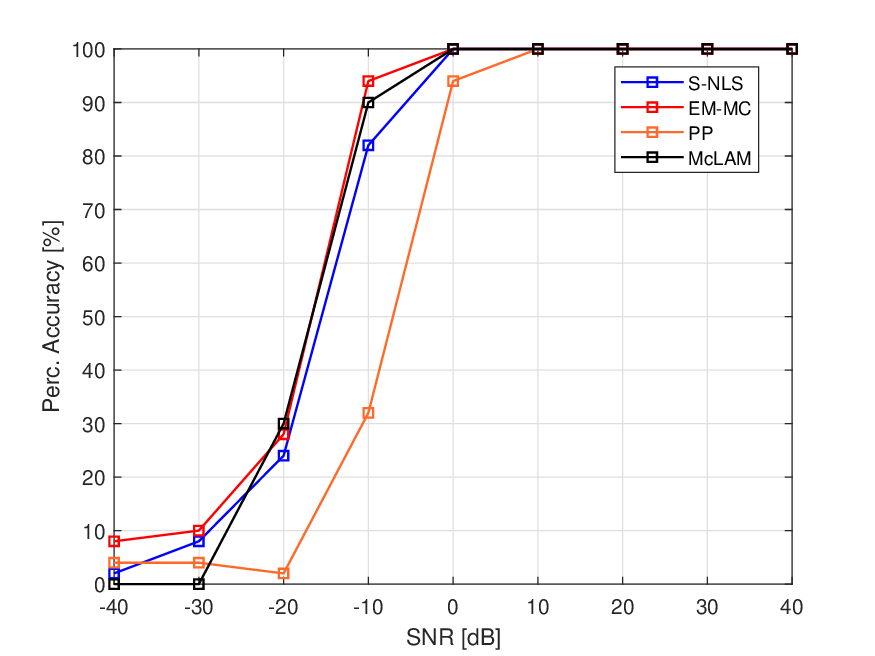}}}
    \qquad
    \subfloat[]{{\includegraphics[scale=0.5]{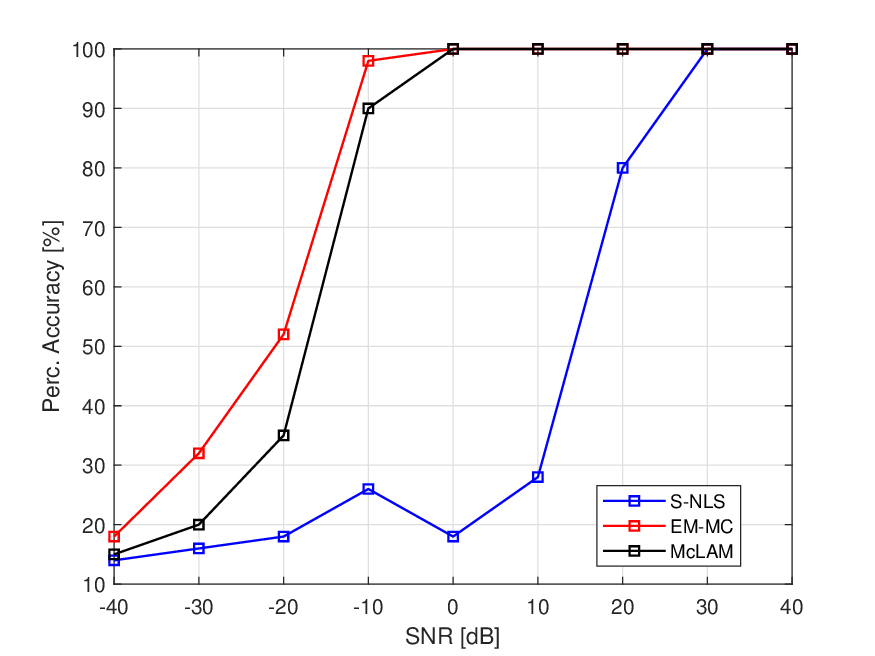}}}
        \caption{a) TOA and b) DOA evaluation under different SNRs}%
    \label{fig:evalToaDoa}%
    \vspace{-0.5cm}
\end{figure*}
\begin{figure*}[ht]
    \centering
   \subfloat[]{{\includegraphics[scale=0.5]{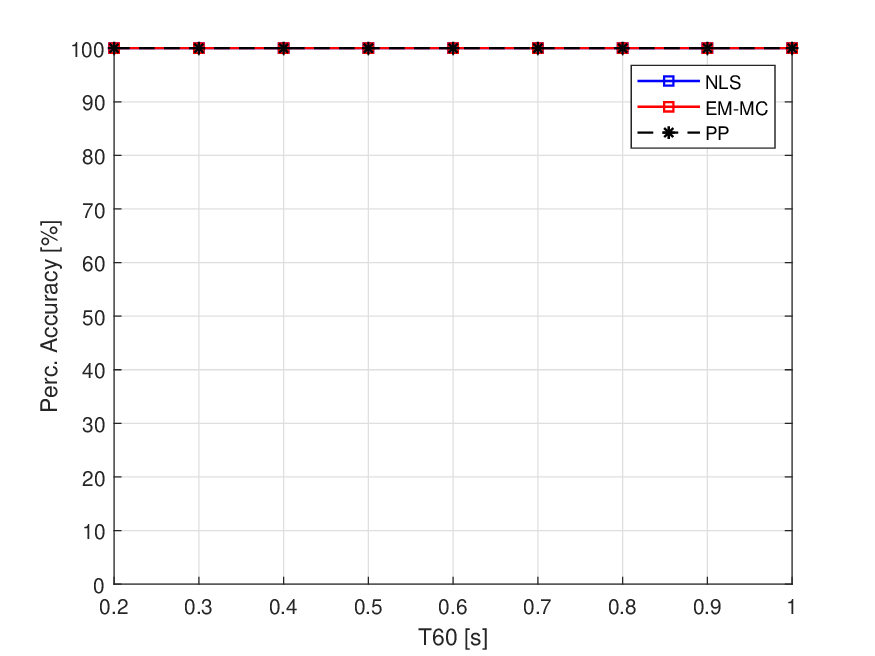}}}
   \qquad
    \subfloat[]{{\includegraphics[scale=0.5]{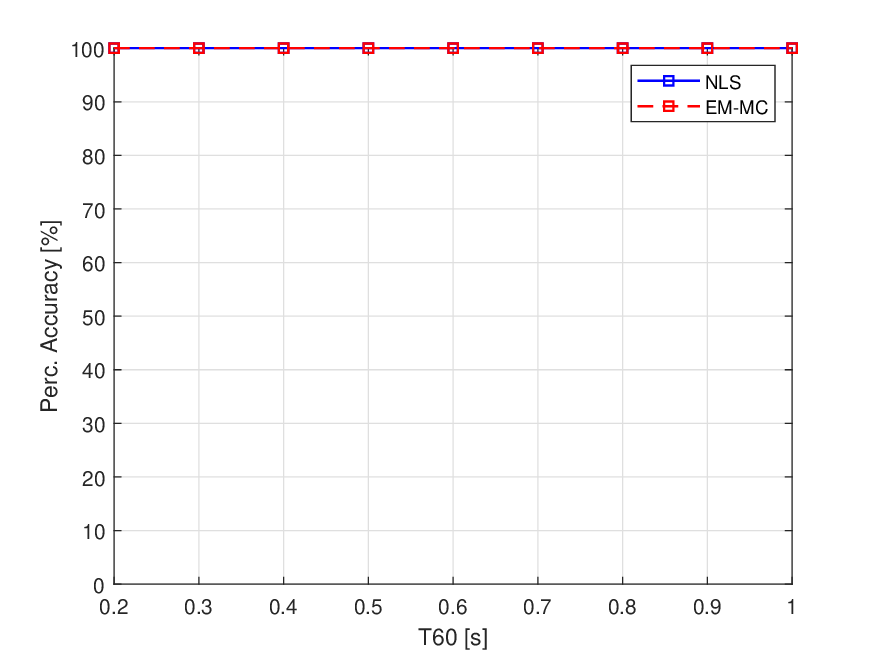}}}
        \qquad
        \caption{a) TOA and b) DOA evaluation under different room reverberation} 
    \label{fig:T60_experiment}%
    \vspace{-0.4cm}
\end{figure*}
\begin{figure*}[ht]
    \centering
   \subfloat[]{{\includegraphics[scale=0.5]{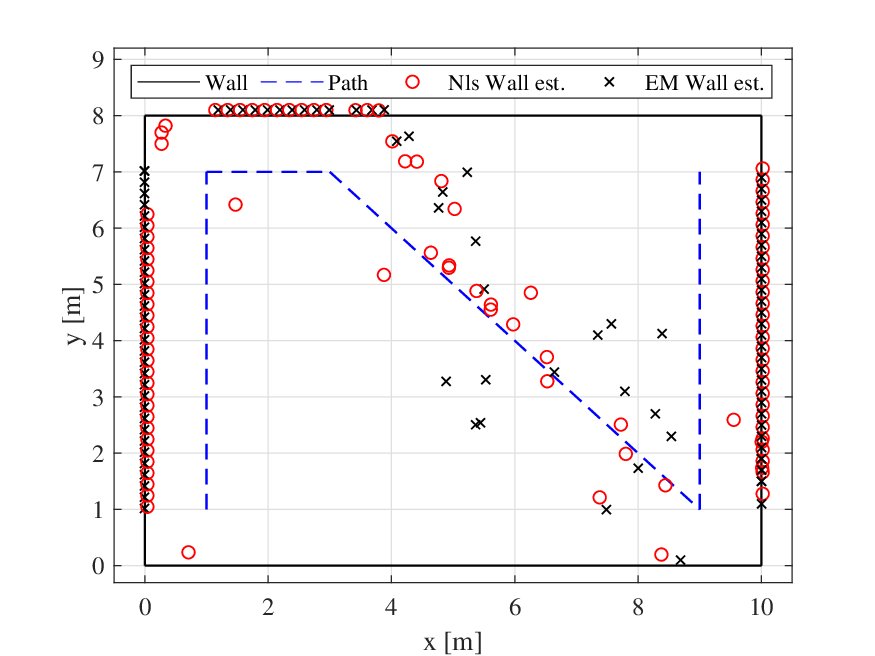}}}
   \qquad
    \subfloat[]{{\includegraphics[scale=0.5]{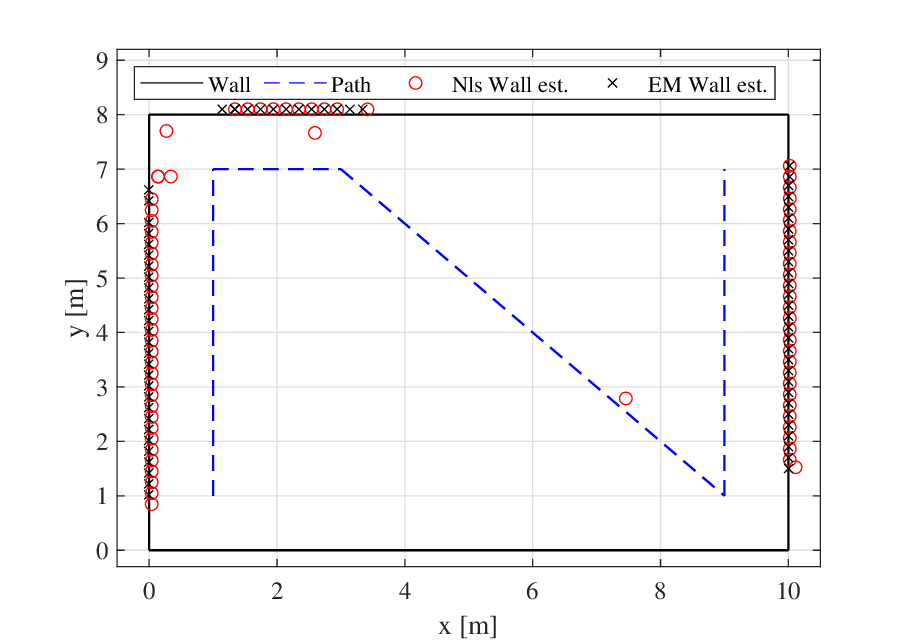}}}
        \qquad
        \caption{Spatial Map of an indoor environment with dimension $8 \times 6 \times 5$ m a) without classifier b) with SVM classifier} 
    \label{fig:acoustic_map_1}%
    \vspace{-0.4cm}
\end{figure*}
\begin{figure*}[ht]
    \centering
   \subfloat[]{{\includegraphics[scale=0.5]{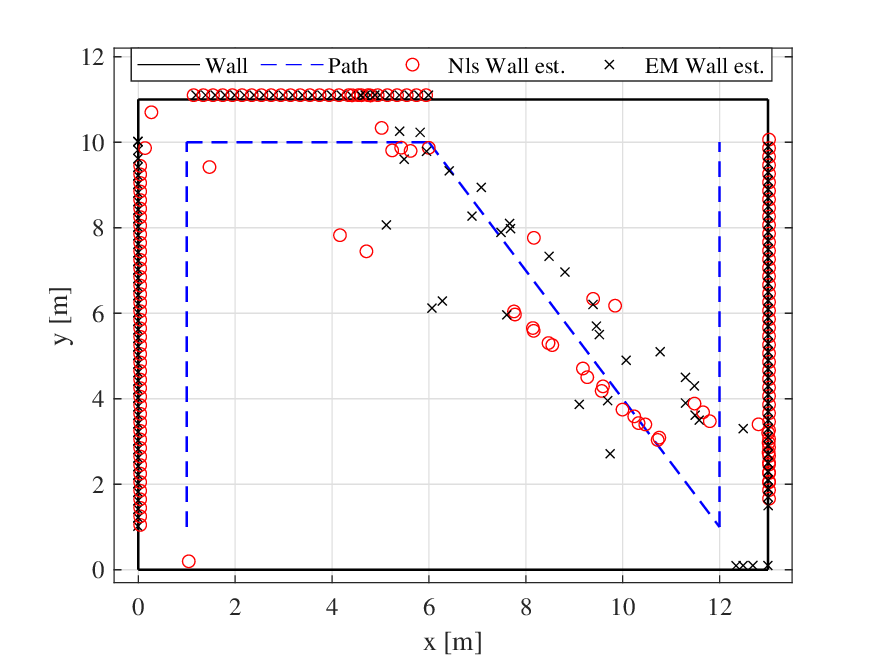}}}
   \qquad
    \subfloat[]{{\includegraphics[scale=0.5]{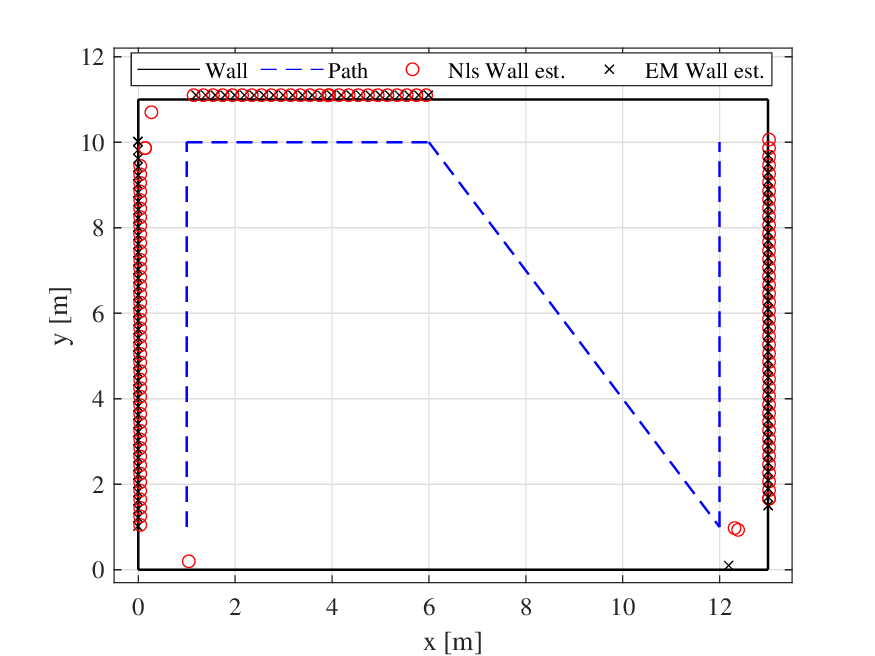}}}
        \qquad
        \caption{Spatial Map of an indoor environment with dimension $13 \times 11 \times 5$ m a) without classifier b) with SVM classifier} 
    \label{fig:acoustic_map_2}%
    \vspace{-0.4cm}
\end{figure*}
\begin{align}
\label{eq:covarianceMatrix}
\mathbf{R}_{y}=\frac{1}{T}\sum_{t=0}^{T-1}\mathbf{Y}(t,\omega)\mathbf{Y}^{H}(t,\omega).
\end{align}
If we assume the signal covariance to be white Gaussian then the MPDR beamformer will reduce to the Delay-And-Sum Beamformer (DSB), which is aligned with the interpretations to the M-step of the EM algorithm in \cite{jensen2019, saqib2020estimation}. Moreover, we regularize the covariance matrix of the observed signal as in \cite{benesty2009}
\begin{equation}
\label{eq:regularizedCovarianceMatrix}
    \overline{\mathbf{R}}_{y} = (1 - \gamma)\mathbf{R}_{y} + \gamma \frac{\text{Tr}\lbrace{\mathbf{R}_{y}}\rbrace{\mathbf{I}}}{M}
\end{equation}
where $\gamma$ is the regularization parameter, and $\text{Tr}(\cdot)$ is the trace of a matrix, and $\mathbf{I}$ is the $M\times M$ identity matrix. The regularization is added to overcome signal cancellation due to estimation errors and reverberation. When evaluating the performance of our estimator, a value of $\gamma = 0.1$ was selected for the experiments. The noise covariance matrix, $\mathbf{R}_{y}$, in \eqref{eq:covarianceMatrix} is then replaced by regularized noise variance matrix \eqref{eq:regularizedCovarianceMatrix}, $\overline{\mathbf{R}}_{y}$.

While these estimators are accurate even under noisy conditions as we will show in the experimental results, they exhibit a thresholding behaviour, like any estimator, when the noise becomes too dominant, e.g., when the platform is too far away from an acoustic reflector. Yet, the estimator will still provide estimates in such conditions, which may lead to spurious estimates in the acoustic reflector mapping.

\subsection{The SVM Classifier}
\label{subsec:trainingSvm}
  The process of acquiring data to train the classifier was done by dividing the room into $1,989$ grid points and placing the robot at each point to probe the environment. The individual classes were labelled such that if the difference between estimated TOAs and the true TOAs were less than 10 samples then the observation was labelled as a "wall" or "0", otherwise it was labelled as "no wall" or "1". A total of $1,989$ observations were spatially synthesized in a room of dimension $10 \times 8\times 7$ m using RIR generator at different positions and SNR values, ranging from $-40$ to $40$ dB. The observations were then divided into training set and test set using a $80:20$ ratio, respectively. A total of $1,591$ observations were used for training while $398$ observations were used for testing. To prevent over-fitting, we used a cross-validation model with $5$ k-fold of equal size. The training was done with a direct-path component, R=1, removed from the observation because the array geometry is known which enables either offline measurement of the impulse response of the direct-path component or analytical computation of the impulse response of the direct-path component.

\section{Experimental Evaluation}

\subsection{Experimental Setup}
\label{subsec: expResults}
The evaluations of the proposed solution are carried out in two simulated environments using the Room Impulse Response Generator \cite{rirGenerator2010}. Although, a room of size $10 \times 8 \times 6$ m was used to train the classifier, a room of size $8\times 6\times 5$ m and a room size of $13 \times 11 \times 5$ m were used to test the proposed framework. For this experiment, $M = 6$ microphones were employed in a UCA of radius $d = 0.2$ m while the source was positioned at the center of the UCA. In the first evaluation, the reverberation time was set to $T_{60}=0.6$ s, while the speed of sound was set to $343$ m/s and the sampling frequency was set to $22.05$ kHz. The transmitted signal $s(n)$ used in the simulation was an additive white Gaussian noise (AWGN) signal constituted by $1,500$ samples drawn from a Gaussian distribution with zero padding to form a signal of length $20,000$. Furthermore, the background noise $v_{m}(n)$ was composed of two parts: diffuse cylindrical noise to model the rotor noise, $f1$, and the background noise of the environment, $f2$, such that $ y = x + f1 + f2$. The diffuse cylindrical noise was generated using the method described in \cite{habet2008} from a sound file downloaded from the DREGON dataset in \cite{dregon2018}, containing rotor noise at a speed of $70$ rotations per second (RPS) while the background noise was generated as spatially and spectrally white noise. The rotor noise in DREGON dataset was sampled at $44.1$ KHz but a down sampling was applied to reduce computational cost. The Signal-to-Diffuse Noise Ratio (SDNR) is defined as ratio between the variance of the observed probe signal at the microphone $1$, $\sigma_{x}^{2}$, against the variance of the rotor noise $\sigma_{f1}^{2}$. The Signal-to-Noise Ratio (SNR) is defined as ratio between the variance of the observed probe signal at the microphone $m$, $\sigma_{x}^{2}$, against the variance of the background noise $\sigma_{f2}^{2}$. In the following subsection, we perform two experiments: A quantitative and a qualitative experiment. Moreover, we adjust the TOA interval search to exclude the direct-path component which resides with UCA size of $0.2$m. Therefore, in the experiments, we define the search interval from $\tau_{min} = 1$m to $\tau_{max}=2$m which will ensure that only the acoustic echo belonging to a wall are estimated.

\subsection{Results}
In the first experiment, the performance of the proposed method was evaluated over several SNR values while the diffuse noise was kept fixed at $40$ dB. In the second experiment, the performance of the proposed method was evaluated over different reverberation level (T60). The accuracy in both experiments are defined as the percentage of TOAs and DOAs that are within $\pm 5$ sample of one of the true TOAs and DOAs. Furthermore, the proposed method was compared against current state-of-the-art such as the multichannel Expectation-Maximization (EM-MC) method \cite{saqib2020estimation}, the multichannel localization and mapping (McLAM) method \cite{saqib2022} and the commonly used peak-picking approach from the estimated RIR. For the later method, the RIR was estimated using dual channel analysis as in \cite{herlufsen1985}. The room dimension for both evaluation was of size $10 \times 8 \times 5$ m.

The proposed method provide good TOA performance of around $80\%$ at SNR of $-10$ dB which is similar to EM-MC, as shown in Fig. \ref{fig:evalToaDoa}(a), Furthermore, as seen in the Fig. \ref{fig:evalToaDoa}(b), the DOA performance of the propose method is good only at higher SNRs. This is because only one microphone was used to estimate $\tau$ of the acoustic echo compared to EM-MC but this provides a decrease in computational load while sacrificing DOA accuracy. Furthermore, the computation time of the RIR-PP, EM-MC, McLAM  and the proposed S-NLS, were measured using MATLAB's built-in function \textit{timeit} on a standard desktop computer running a Microsoft Windows 10 operating system with an Intel Core i7 CPU with 3.40 GHz processing speed and 16 GB of RAM. A Monte Carlo simulation with 50 trials was performed on each method and an average time was calculated. The measured computation times of the RIR-PP, EM-UCA and the proposed method are $0.0063$s $63.25$s, $60.65$s and $10,14$s, respectively, for R=1 and an SNR of 40 dB. Both plots in Fig. \ref{fig:evalToaDoa}(a) and (b) were measured over $50$ Monte-Carlo simulations.

In the second experiment, the performance of the proposed method was tested under different reverberation level (T60) in a simulated environment. As seen in Fig. \ref{fig:T60_experiment}, both TOA and DOA estimationn of the proposed method is unaffected under different reverberation level from $[0.2 - 1]$ s.

\subsection{Application Example}
In this section two experiments were performed to test the framework. The experiments were performed in a simulated room of dimension $8\times 6\times 5$ m and a room size of $13 \times 11 \times 5$ m. In these experiments, the robot takes an elaborate trajectory as shown in Fig. \ref{fig:acoustic_map_1} and Fig. \ref{fig:acoustic_map_2}. The first test was conducted without a classifier and with direct-path component removed from the observed signal. As shown in Fig.\ref{fig:acoustic_map_1}(a) and Fig. \ref{fig:acoustic_map_2}(a), spurious estimates at empty space are seen in the absence of a SVM classifier. However, in Fig. \ref{fig:acoustic_map_1}(b) and Fig.\ref{fig:acoustic_map_2}(b), the tests were performed with SVM classifier, which reduces the number of estimates that belongs to an empty space. 

\section{Discussion and Conclusion}
\label{sec: conclusion}
In this paper, we address the problem of generating a spatial map of an indoor environment using echolocation alone on a robot. We proposed a framework that utilizes a S-NLS estimator and a DOA estimator which are subsequently linked to a SVM classifier to remove spurious estimates. The advantage of the proposed method over traditional sensing technologies such as lidar and camera-based techniques is that the proposed method could be used for resource constraint robots that work in dark environment and detect transparent surfaces. As seen in Fig.\ref{fig:evalToaDoa}, the proposed TOA estimator method is comparable to the EM-MC, McLAM methods and supersedes the peak-picking approach at low SNR of $-10$~dB. However, S-NLS offers low DOA accuracy which is a sacrifice we make for reduce computational load of $10$s. Furthermore, in the qualitative experiment, we see that combining the TOA/DOA estimators with SVM classifier is useful in removing spurious estimates for efficient spatial map generation. One major advantage of using SVM classifier is that it can be trained with additional parameters to estimate the robot's position within a room. In the future iteration of this work, we aim to implement the proposed method on an actual robot for spatial map generation and test the performance of the proposed method in both indoor and outdoor environment as well as improve the DOA accuracy by pre-whitening the observed signals.

\addtolength{\textheight}{-12cm}   




\bibliographystyle{IEEEtran}

\end{document}